\documentclass{SIP} 

\usepackage{achemso}
\usepackage{multirow}

\usepackage{epstopdf, epsfig} 
\usepackage[export]{adjustbox}

\DeclareMathOperator*{\argmax}{\arg\max}
\DeclareMathOperator*{\argmin}{\arg\min}

\begin{document}

\title{Deep Active Learning for Computer Vision: Past and Future}

\author[Rinyoichi Takezoe, \textit{et al}.]{Rinyoichi Takezoe$^{1, 2}$, Xu Liu $^{3}$, Shunan Mao $^{2}$, Marco Tianyu Chen $^{4}$, Zhanpeng Feng $^{1}$,  Shiliang Zhang $^{2}$, Xiaoyu Wang $^{1}$}

\address{
\add{1}{Intellifusion Inc.}
\add{2}{Peking University}
\add{3}{National University of Singapore}
\add{4}{Shenzhen Institute of Advanced Technology, Chinese Academy of Sciences}
\add{*}{The first four authors contributed equally to the paper.}
}



\begin{abstract}
As an important data selection schema, active learning emerges as the essential component when iterating an Artificial Intelligence (AI) model. It becomes even more critical given the dominance of deep neural network based models, which are composed of a large number of parameters and data hungry, in application. Despite its indispensable role for developing AI models, research on active learning is not as intensive as other research directions. In this paper, we present a review of active learning through deep active learning approaches from the following perspectives: 1) technical advancements in active learning, 2) applications of active learning in computer vision, 3) industrial systems leveraging or with potential to leverage active learning for data iteration, 4) current limitations and future research directions. We expect this paper to clarify the significance of active learning in a modern AI model manufacturing process and to bring additional research attention to active learning. By addressing data automation challenges and coping with automated machine learning systems, active learning will facilitate democratization of AI technologies by boosting model production at scale.
\end{abstract}



\maketitle
\section{Introduction}
To democratize Artificial Intelligence (AI) for all industry verticals, researchers have been working on automating the machine learning design process commonly known as AutoML. Through techniques like neural network architecture search \cite{liu2018progressive,tan2019mnasnet}, automated hyper-parameter optimization \cite{akiba2019optuna,falkner2018bohb}, meta-learning \cite{finn2019online,finn2017model}, etc., AutoML eases the requirements of domain knowledge and facilitates the building of high-quality machine learning models for developers from academia and industry.

Different from academia where the common practice is to use a fixed and pre-defined dataset for model iteration, the life cycle of developing an industrial model involves both model iteration and data iteration. Advances in deep learning have made it such that even standard deep models without careful design can perform decently. In contrast to the breadth of works advancing deep learning for developing sophisticated application-specific models, the investment in data iteration technologies is limited. However, robust data is pivotal to developing strong real-world models. The dataset largely determines the model's precision upper bound as well as its generalization capability when applied to novel scenes. In some scenarios, collecting more data is equally important, comparing with employing a more powerful machine learning algorithm. Therefore, the traditional paradigm of machine learning model development is arguably shifting from Model-centric to Data-centric approaches~\cite{data-centric21}. In our technology development process, a vast majority of efforts were spent on the curation of the right data, comparing to inventing better models when conducting AI model production. Therefore, automating or accelerating the data iteration process is the key for rapidly developing AI solutions.

In this paper, we focus on the process of data selection automation. Active learning (AL) is one of the core data selection automation technologies and has gained traction in recent decades. The goal of AL is to alleviate the expensive data labeling process and construct a resource-efficient yet robust training set by selecting the most valuable samples for human annotation. Specifically, AL operates in an iterative fashion. At each step, AL algorithms select and annotate informative samples from an unlabeled data pool, which are then used to train and update machine learning models. Although only a subset of the total data may be used, research suggests that this resource-efficient AL approach can still yield desired results. For example, models using AL identified data can use only 40\% of the training data to achieve performance on par with that of using all data \cite{caramalau2021sequential}. Furthermore, the improvement of AL may be even more significant in real-world industrial applications where data come from open-world scenarios and contain unavoidable noise, necessitating costly professional involvement.

Active learning methodologies are also evolving as deep learning is paving its way for all different kinds of applications. With millions or billions of parameters, training deep models requires much more data compared to that of traditional machine learning algorithms, which makes efficient data selection technology increasingly critical. Deep Active Learning (DeepAL), an approach combining DL and AL to maximize task performance while minimizing the cost of data labeling, has attracted attention and shown great promise for various tasks such as image classification \cite{zhang2020state,yoo2019learning}, object detection \cite{sener2018active,roy2018deep}, text classification \cite{an2018deep,zhang2017active}, and sentiment analysis \cite{bhattacharjee2017active,zhou2010active}.

There exist some efforts reviewing active learning related literature. \citet{ren2021survey} conducted a broad literature review for DeepAL in general AI applications and included relevant developments until 2020. In contrast, \citet{wu2020multi} provided a thorough survey concentrated on multi-label active learning for image classification tasks. More recently, \citet{zhan2022comparative} delivered a comparative survey with an emphasis on the performance comparison of different DeepAL methods under varying tasks. While existing surveys either discuss active learning from the perspective of broad AI research including vision, nature language processing and machine learning, or dive deep into a specific vision task, there lacks a survey reviewing broader vision-based DeepAL methods. Moreover, active learning has potential to significantly accelerate the data iteration or annotation in industrial system. However, there is no discussion about how active learning is integrated into modern deep model manufacturing software and machine learning operations.

This paper provides a comprehensive survey about deep active learning in the context of computer vision from both academia and industry perspectives. Existing deep active learning research is far from mature and its application in practical systems is far from well explored. Through reviewing current technique developments, explaining and illustrating their applications in advanced artificial intelligence systems, and discussing unresolved problems, we hope this paper will serve as a succinct and relevant summary for the community.

The rest of the paper is organized as following. First, we explain the basics of AL and present an overview of recent developments in DeepAL methods. Second, we discuss the usage of DeepAL algorithms for computer vision tasks, including image recognition, object detection, semantic segmentation, and video recognition. Third, we provide an introduction to the practical applications of DeepAL methods in industry software or systems. Lastly, to promote the prosperity of DeepAL, we discuss current limitations and potential directions in this field.

\section{Overview of Active Learning}
\label{overview}

\begin{figure}[ht]
  \centering
  \includegraphics[width=\linewidth]{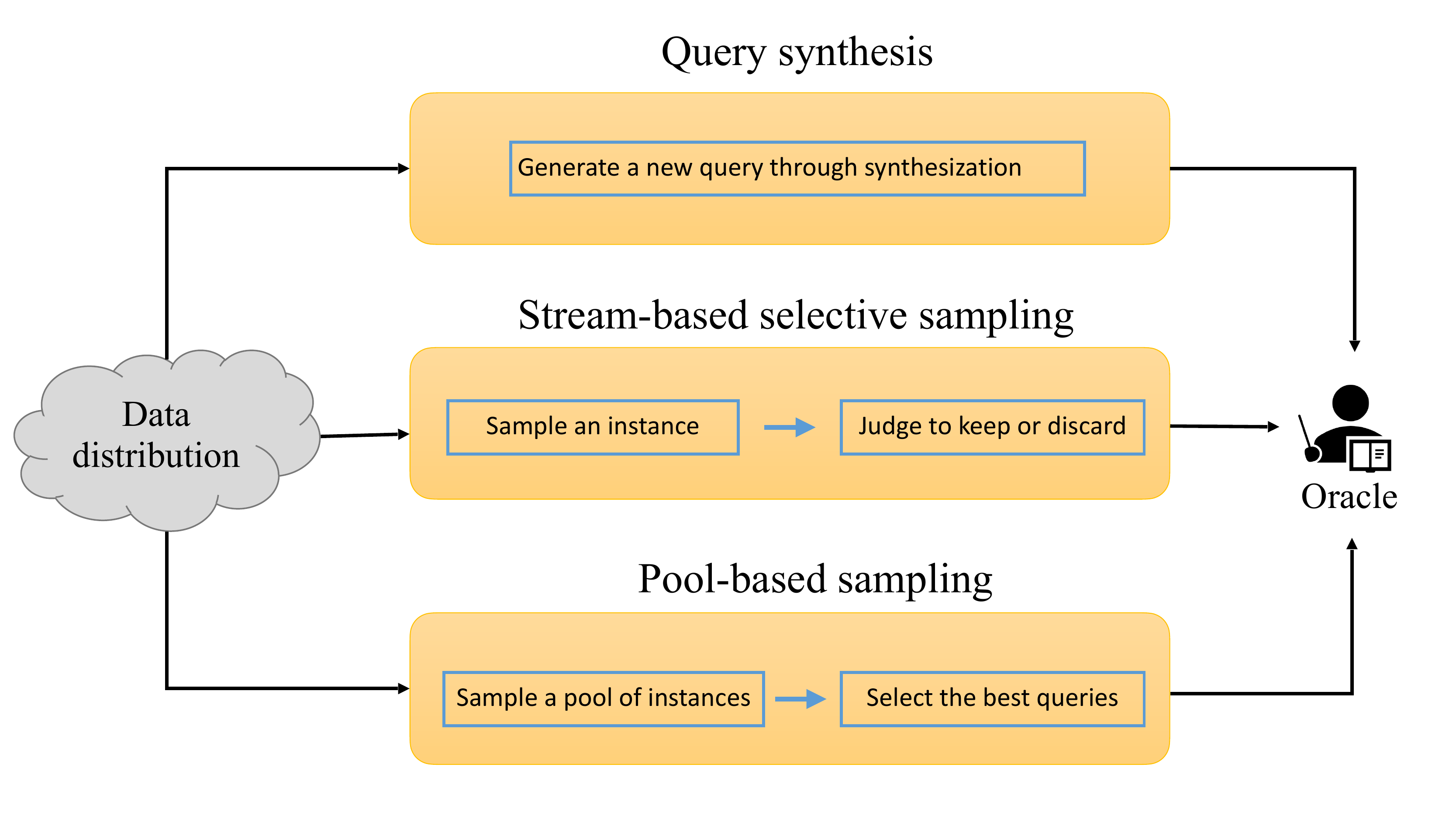}
  \caption{Three scenarios of active learning. 'Pool-based Sampling' is used in most DeepAL settings.}
  \label{fig:scenario}
\end{figure}

\subsection{What is Active Learning?}
Active learning is the task of choosing the most valuable data for a learning algorithm so that it can perform similarly or even better with the resulting less training data. More practically, an active learning system selects the most valuable unlabeled samples by a query strategy, then sends them to human for labeling or curation. The labeled data is then adopted for next stage model training.

There are arguably three commonly studied active learning scenarios in existing literature, including stream-based selective sampling, membership query synthesis, and pool-based sampling, respectively as shown in Fig.~\ref{fig:scenario}.
Stream-based selective sampling processes every unlabeled sample independently, so the resulting number of samples for labeling is not fixed. This sampling rule is suitable for sequentially provided online data, \eg~Robot-Assisted AL~\cite{suzuki2021annotation}. In this scenario, it is hard to calibrate the threshold to determine whether a sample should be selected or not.
Membership query synthesis generates new data to be annotated. It is efficient in the interpretable feature space, but may suffer from low-quality generation because of the limited knowledge of data distribution in unseen domains.
Pool-based sampling chooses the best query samples among the entire unlabeled set. It commonly computes a score function for each unlabeled sample and makes the selection according to the score, \eg, samples with the highest scores are sent for annotation. The vast majority of DeepAL methods in the literature use pool-based sampling and they try to select the most valuable data from the unlabeled pool. Some DeepAL methods adopt membership query synthesis with deep image generation methods.


 


\subsection{Deep Learning-based Methods}

\subsubsection{Generic Deep Active Learning Framework}
The advent of deep learning has made DeepAL a popular topic. The primary characteristic of DeepAL is that it heavily relies on batch-based sample querying.
Traditional per-sample query methods lead to frequent model retraining. Since each retraining of the model is computationally expensive, the one-by-one query method is inefficient for deep learning. Batch mode query strategies are frequently employed in DeepAL to strike a balance between sampling (for annotation) cost and training cost. In a cycle of active learning, the first step is to get the optimal deep model using existing labeled data. Then, the most valuable batch from the unlabeled pool is selected using active learning based on features extracted by the resulting deep model. The performance of the deep model is expected to converge after several rounds of model training and active learning.

In each cycle, one can choose to retrain the deep model or just perform fine-tuning solely using newly collected data. While retraining a model could be very costly, fine-tuning a model could cause divergence or drifting from the original model.
Except for a few works, \eg~CEAL~\cite{wang2016cost}, which fine-tune the model, most deep active learning methods retrain the deep model from scratch at each active learning cycle, to ensure the model performance is not compromised. As the fast growth of GPU computation power, the cost of training a deep model is also decreasing. The number of active cycles depends on the computation budget and time budget of a project. A project could last for weeks or years.

A generic deep active learning framework can be formulated as follows.
Assuming the whole active training procedure consists of $T$ cycles, where each cycle is further divided into two steps: the exploitation step and the exploration step. 
The exploitation step trains a deep model $f(\cdot;\theta^t)$ from the initial state with labeled samples $D_l = \lbrace(x_i,y_i)\rbrace_{i=1}^M$, where $\theta^t$ are the parameters of the deep model to be optimized in the $t$-th cycle.
The exploration step selects a subset of unlabeled samples for labeling. First, each sample from the unlabeled pool $x\in D_u = \lbrace(x_j)\rbrace_{j=1}^N$ receives an informativeness score denoted as  $\operatorname{Score}(x,\theta)$. It is obtained based on extracted sample features or predictions by the deep model with the parameters $\theta$. Then, a batch of samples with the highest scores are selected and sent to the labeling experts. Finally, the labeled batch joins the existing labeled set to formulate a new training set.
Therefore, the $t$-th cycle in a typical active learning process can be formulated as
\begin{equation}\label{eq:BMAL}
\begin{aligned}
&\theta^t&&=\argmin_{\theta}\sum_{\{x,y\}\in D_l^t}\mathcal{L}(f(x;\theta),y)\\
&\mathcal{B}^t&&=\argmax_{\mathcal{B}\subset D_u^t}\sum_{x\in \mathcal{B}} \operatorname{Score}(x,\theta^t)\\
&D_l^{t+1}&&=D_l^{t} \cup \mathcal{B}^t,
\end{aligned}
\end{equation}

where $D_l^t$ and $D_u^t$ denote the labeled set and the unlabeled pool in the $t$-th cycle. $\mathcal{L}$ is the loss function employed to train the deep model $f(\cdot;\theta)$, and $\operatorname{Score}(\cdot)$ is the function measuring the information obtained from each unlabeled sample.

\begin{figure}[t]
  \centering
  \includegraphics[width=\linewidth]{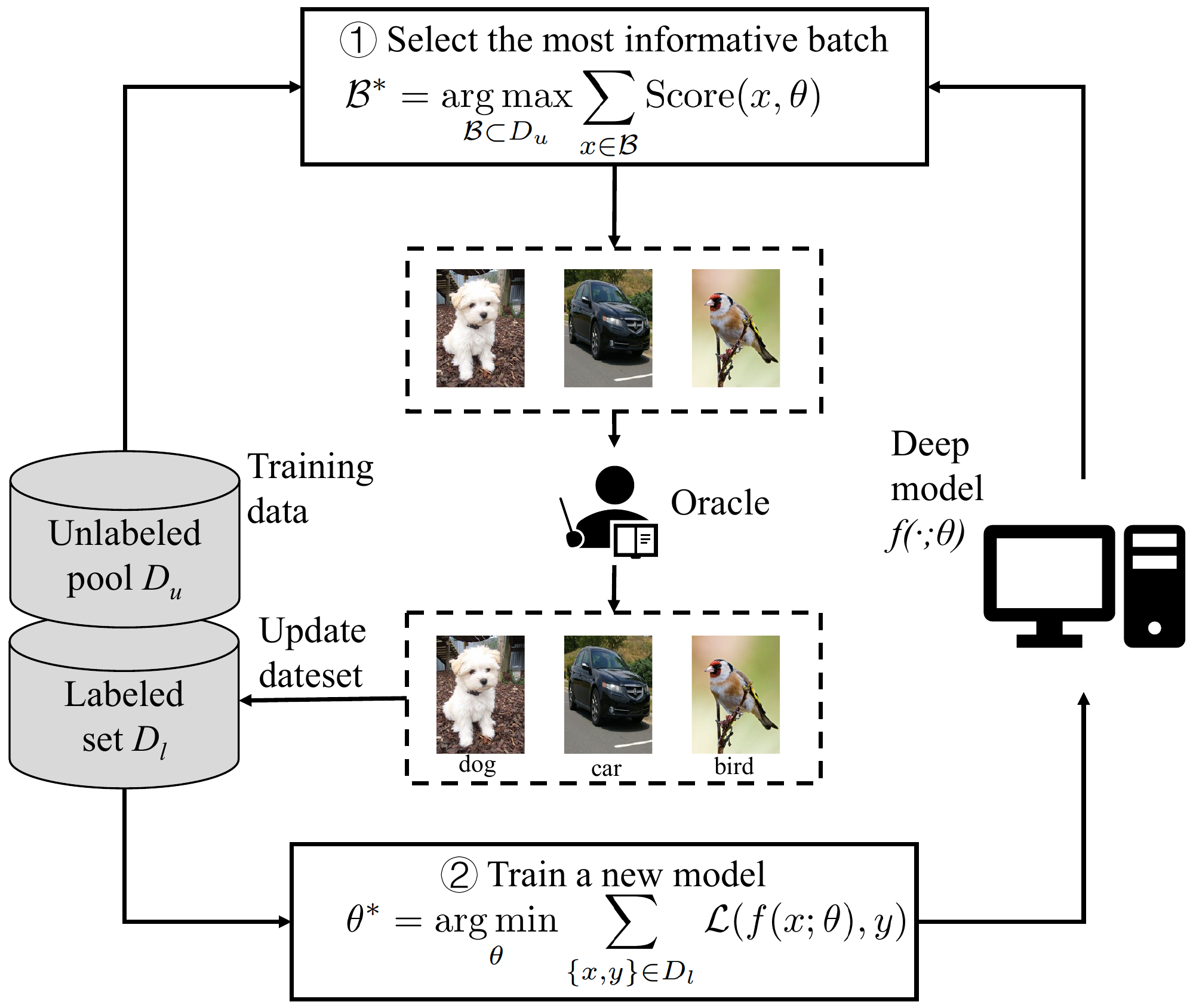}
  \caption{A typical deep active learning framework for vision tasks.}
  \label{fig:genericframe}
\end{figure}

Most existing DeepAL research focuses on solving the following two problems: 1) selecting the most valuable batch and 2) training a deep model with insufficient labeled data. Related methods will be discussed in the following two parts. 

\subsubsection{Selecting the Most Valuable Batch}
DeepAL aims to achieve high performance with limited annotation budgets. To make full use of the annotation budget, it is critical to select a batch of informative samples to maximize performance gain. To achieve this goal, different methods have been proposed to measure the informativeness of each sample. These methods mainly identify informative samples by two criteria: samples that the model is uncertain about and samples that have a different feature distribution with respect to already labeled data. Some works also combine these two criteria to perform sample selection.

\textbf{Model Uncertainty-based Approaches:} The uncertain samples can be regarded as hard-to-learn samples for the current model. Using uncertain samples in the training loop can let the model focus on the task that it is not good at, instead of keep learning from simple samples. The uncertainty can be evaluated using the model's predictions. The following part reviews classic methods that are widely used in traditional AL.

Least Confidence (LeastConf)~\cite{lewis1994heterogeneous} selects data points whose predicted labels \^{y} have the lowest posterior probability, \ie,
\begin{equation}\label{eq:Least Confidence}
\begin{split}
\operatorname{Score}(x,\theta)=-p(\hat{y}|x,\theta),
\end{split}
\end{equation}
where $\theta$ denotes the model parameters, $p(y|x,\theta)$ is the posterior probability that the sample $x$ is predicted to class $y$ by $\theta$.

Margin~\cite{roth2006margin} selects data points, where the difference between the highest predicted probability and the second-highest predicted probability is small. The corresponding score function can be denoted as,
\begin{equation}\label{eq:Margin sampling}
\begin{split}
\operatorname{Score}(x,\theta)=-[p(\hat{y_1} |x,\theta)-p(\hat{y_2}|x,\theta)],
\end{split}
\end{equation}
where $\hat{y_1}$ and $\hat{y_2}$ are the first and second predictions with the highest probability. A small difference indicates that the model is not able to distinguish one class from the other.

Entropy~\cite{joshi2009multi} selects data points that maximize the predictive entropy:
\begin{equation}\label{eq:Maximum Entropy}
\begin{split}
\operatorname{Score}(x,\theta)&=\mathbb{H}[y|x,\theta]\\
&=-\sum_{c}p(y_c|x,\theta)\log p(y_c|x,\theta).
\end{split}
\end{equation}
Compared with LeastConf and Margin, which only consider predictions with the highest probability, Entropy utilizes the whole posterior probability distribution to measure the uncertainty. A larger entropy hence suggests a higher sample uncertainty.

Bayesian Active Learning by Disagreements (BALD)~\cite{houlsby2011bayesian} aims to minimize the uncertainty of parameters. It can be approximated as selecting data points expected to maximize the conditional mutual information between model predictions and model parameters. The uncertainty score can be denoted as,
\begin{equation}\label{eq:BALD}
\begin{split}
\operatorname{Score}(x,\theta)&=\mathbb{I}[y,\theta|x,D_l]\\
&=\mathbb{H}[y|x,D_l]-\mathbb{E}_{p(\theta|D_l)}[\mathbb{H}[y|x,\theta]].
\end{split}
\end{equation}
Maximizing Eq.~\eqref{eq:BALD} leads to seeking the data points where the model is most uncertain about the average predictions $y$ of the probabilistic model. Meanwhile, predictions of these data points under a certain setting, \eg data augmentation, are confident.

Mean Standard Deviation (Mean STD)~\cite{kampffmeyer2016semantic} measures the model uncertainty by the mean standard deviation of the posterior probabilities over all classes:
\begin{equation}\label{eq:Mean STD}
\begin{split}
\operatorname{Score}(x,\theta)&=\frac{1}{C}\sum_{c}\sqrt{Var_{q(\theta)}[p(y_c|x,\theta)]}
\end{split}
\end{equation}

Most of these score functions perform well in traditional active learning, but can not extended to DeepAL. One of the reasons is that many score functions rely on model prediction, which easily become overconfident due to the characteristic of the softmax layer in a CNN model. In order to adapt the above methods to the DeepAL, \citet{gal2017deep} proposes Deep Bayesian Active Learning (DBAL) that combines Bayesian Convolutional Neural Networks (Bayesian CNNs) into the active learning framework. Bayesian CNNs perform variational inference to approximate the true model posterior, which provides model uncertainty cues that can be used in the score functions. In practice, DBAL makes use of Monte-Carlo dropout (MC dropout) stochastic regularisation technique to perform approximate inference. 

Similarly, another work~\cite{beluch2018power} proposes Ensembles which uses an ensemble of CNNs to estimate prediction uncertainty. Specifically, Ensembles trains an ensemble of $N$ classifiers with the same $D_l$ and same network architecture, but with different random weight initializations. It uses the average softmax vectors of $N$ classifiers as the output. The score functions can be applied to Ensembles as well. Ensembles generally perform better than DBAL due to the larger model capacity and the diversity of ensemble models.

In addition to estimating the uncertainty from outputs statistically, there also exist some methods that estimate the uncertainty from other perspectives. \citet{ducoffe2018adversarial} argue that commonly-used methods might be overconfident since a small modification on a sample can lead to an unexpected misclassification. They, therefore, propose Deep-Fool based Active learning (DFAL) to select data points near the decision boundary. DFAL approximates the distance between a sample and the decision boundary by the distance between this sample and its nearest adversarial sample. The selected and adversarial samples are added to the training set.

Another kind of method tries to generate samples with the highest uncertainty instead of selecting from available samples.
They are more similar to active learning by query synthesis than standard pool-based active learning. \citet{zhu2017generative} propose Generative Adversarial Active Learning (GAAL) which combines GAN and active learning. They generate query samples with the largest uncertainty to get good performance with the least number of labeled data. \citet{tran2019bayesian} introduce more data generation methods like Bayesian data augmentation to generate samples in the intersection of different classes. 

According to the proposed influence function by \citet{koh2017understanding}, \citet{wang2022boosting} have theoretically proven that selecting data points of higher gradient norm leads to a lower upper-bound of test loss, resulting in better test performance. However, the loss cannot be computed with a lack of labeled data. To address this issue, \citet{ash2019deep} proposes Batch Active learning by Diverse Gradient Embeddings (BADGE) that assigns pseudo labels based on model predictions, and \citet{wang2022boosting} propose two schemes named expected-gradnorm and entropy-gradnorm, respectively. These schemes calculate expected loss and entropy as a proxy. 

Since computing gradient norms needs to estimate the loss first, \citet{yoo2019learning} propose a task-agnostic method that directly predicts loss to measure the uncertainty. This method is called Learning Loss for Active Learning (LL4AL). LL4AL attaches a small parametric loss prediction module to the task model, and jointly trains both models. The loss prediction module is trained to predict the losses of the task model for unlabeled samples. Samples with high loss will be selected for annotation and training. 

\textbf{Feature Distribution-based Approaches:}
Another category of methods selects samples that are different from labeled ones in the feature space. Once a sample is labeled, it can be regarded as a representative of the feature distribution. This makes it no longer necessary to annotate similar samples. The key intuition is that if selected features are able to represent the true data distribution, the model trained with AL will perform similarly to the model trained with the full data.

A straightforward method is to utilize the distance or similarity between data points in the feature space and avoid selecting similar samples. 
\citet{yin2017deep} propose a similarity-based method consisting of two stages: exploitation and exploration. The exploitation stage selects samples with maximum uncertainty and minimum redundancy, where the redundancy is measured by the similarity within the selected set. The exploration stage selects samples farthest from the labeled set to find diverse samples. 
\citet{biyik2019batch} employ k-Determinantal Point Processes (k-DPPs), a class of repulsive point processes that select a batch of samples with probability proportional, to the determinant of their similarity matrix.

\citet{geifman2017deep} and \citet{sener2018active} define the problem as core-set selection, where the goal is to find a subset such that a model learned on it can achieve comparable performance with learning on the entire dataset. \citet{sener2018active} theoretically prove that the objective is equivalent to the k-center clustering problem. Since the problem is NP-Hard,~\citet{sener2018active} use the Farthest-First (FF) traversal to get a greedy approximation in practice. Given a set of unlabeled samples, FF traversal iteratively picks samples farthest from the labeled samples in the feature space generated by the representation layer of the neural network. 

Adversarial learning has also been adopted in AL for querying samples. Specifically, adversarial learning can be conducted between a classifier and a sample selection model. \citet{gissin2019discriminative} propose Discriminative Active Learning (DAL), which poses active learning as a binary classification task between labeled and unlabeled classes. Choosing samples to label in this way makes the labeled and unlabeled data indistinguishable. Similarly, \citet{sinha2019variational} propose Variational Adversarial Active Learning (VAAL), which uses a $\beta$-variational autoencoder ($\beta$-VAE) to map labeled and unlabeled data into the same latent space and trains an adversarial network to distinguish the encoded features. The $\beta$-VAE and the adversarial network are learned together in an adversarial fashion. \citet{shui2020deep} further propose Wasserstein Adversarial Active Learning (WAAL) that adopts Wasserstein distance for distribution matching. They reveal that Wasserstein distance can better capture the diversity.

\textbf{Hybrid Approaches:}
It is reasonable to combine the above two criteria for active learning, \ie, utilizing both model prediction and model features for sample selection. The model prediction is helpful for identifying difficult data points but might ignore the correlation among them. The model feature makes it easier to identify representative data points by considering feature relationships but suffers from degraded effectiveness when the query batch is small. A variety of hybrid strategies have been proposed to leverage both complementary criteria, and have demonstrated better performance. 

A simple hybrid approach is to divide the query procedure into two stages, and different types of samples are selected at each stage.
BADGE~\cite{ash2019deep} computes the gradient embedding for each unlabeled sample in the first stage, then uses the k-means++ seeding algorithm for sampling in the second stage. BADGE tends to select samples with both high diversity and magnitude. \citet{zhdanov2019diverse} first prefilter top $\beta k$ ($\beta \geq 1$) informative samples based on Margin, then cluster them to $k$ clusters with (weighted) k-means. Finally, $k$ samples closest to the cluster center are selected. In contrast, Cluster-Margin~\cite{citovsky2021batch} first performs Hierarchical Agglomerative Clustering (HAC) as a preprocessing step. For subsequent sampling iterations, Cluster-Margin retrieves clusters on which the model is least confident, and uses a round-robin scheme for sampling. Compared with other methods that need to run a diversification algorithm at each sampling iteration, Cluster-Margin enjoys better efficiency because HAC is only executed one time.

Another commonly used approach is to design a unified score function that considers both uncertainty and diversity. \citet{yin2017deep} propose a linear combination of two criteria as the score function for the exploitation step, \ie,
\begin{equation}\label{eq:Exploitation}
\begin{split}
\operatorname{Score}(x,\theta)&=\mathbb{H}[y|x,\theta] - \alpha \operatorname{Sim}(x,S),
\end{split}
\end{equation}
where $\operatorname{Sim}(\cdot)$ computes the similarity between one sample and the selected set, and $\alpha$ denotes a parameter to balance uncertainty and diversity. During the acquisition process, the uncertain samples different from the selected set are chosen one by one. 

\citet{kirsch2019batchbald} extend BALD~\cite{houlsby2011bayesian} to BatchBALD to compute the mutual information between a batch of data points and model parameters, instead of independently computing for each single sample,
\begin{equation}\label{eq:BatchBALD}
\begin{split}
\operatorname{Score}(x_{1:b},\theta)&=\mathbb{I}[y_{1:b},\theta|x_{1:b},D_l]\\
&=\mathbb{H}[y_{1:b},|x_{1:b},D_l]\\
&-\mathbb{E}_{p(\theta|D_l)}[\mathbb{H}[y_{1:b},|x_{1:b},\theta]],
\end{split}
\end{equation}
where $x_{1:b}$ and $y_{1:b}$ denote $b$ selected data points and their predictions. BALD may overestimate the joint mutual information due to double counting of overlaps between data points in the feature space. BatchBALD takes this issue into account and makes itself more likely to acquire a diverse cover.

Those methods need to compute the correlation between samples. This constraint suppresses the capability of  computing representativeness scores individually for each sample. WAAL~\cite{shui2020deep} tackles this issue by using a discriminator to directly estimate the probability that a sample come from the unlabeled pool.

Moreover, compared with explicitly considering the hybrid query strategy, some works aim to balance uncertainty and representativeness in an implicit way. \citet{zhang2020state} propose a State Relabeling Adversarial Active Learning model (SRAAL) which is motivated by the fact that previous works such as VAAL~\cite{sinha2019variational} consider all the unlabeled samples of the same quality. They build the discriminator that maps data to a hard label, \ie, 0 and 1 for unlabeled and labeled data, respectively. To alleviate the issue of hard labels, an online uncertainty indicator in the discriminator is designed to relabel the state information of unlabeled data according to different importance. Task-Aware Variational Adversarial Active Learning (TA-VAAL)~\cite{kim2021task} is another extension based on VAAL~\cite{sinha2019variational}, which combines with LL4AL~\cite{yoo2019learning}. By exploiting RankCGAN~\cite{saquil2018ranking}, the goal of LL4AL is relaxed from predicting the accurate loss value to estimating ranking loss cues.

\subsubsection{Training with Insufficient Data}
Deep learning is data-hungry, and its high performance relies heavily on numerous training data. Standard active learning methods train deep models on the labeled sample set and ignore the unlabeled pool. Therefore, many semi-supervised methods have been proposed to alleviate the issue of insufficient training data.
Semi-supervised learning and active learning are the two sides of the same coin in that, both pursue to achieve good performance with the fewest labeled data. The former focuses on the utilization of unlabeled data, while the latter focuses on selecting and annotating the most valuable data. The combination of semi-supervised learning and active learning has the potential to significantly increase the variety of training data. Semi-supervised learning methods in active learning can be achieved by three means: Pseudo-label, self-supervised learning and data generation.

Pseudo-label methods assign pseudo labels to samples in the unlabeled pool. \citet{wang2016cost} propose CEAL to assign pseudo labels according to model predictions. They progressively select two types of samples from the unlabeled set to fine-tune the deep model. Samples with high uncertainty are selected for labeling. Samples with high confidence are assigned pseudo labels without human cost. \citet{simeoni2021rethinking} propose a transductive way to generate pseudo labels. Following \citet{iscen2019label}, the prediction of the unlabeled data is made according to the label propagation on their nearest neighbors. Model predictions of the unlabeled data are used to compute the certainty of pseudo labels. \citet{gorriz2017cost} propose a similar cost-efficient method to leverage unlabeled samples in active medical segmentation. They reduce the uncertainty map to a single uncertainty value for active sample selection and pseudo label assignment. Samples with the lowest uncertainty are added to the training set along with their predicted masks.

Self-supervised methods utilize the unlabeled samples for training.
\citet{liu2016active} propose an unsupervised feature learning on all datasets to improve the training efficiency of Deep Belief Network (DBN). \citet{gao2020consistency} propose a consistency-based unsupervised training to learn from unlabeled data by optimizing the feature distance between unlabeled samples and their augmentations. In addition, they also propose a consistency-based sample selection strategy to choose augmented samples with the largest inconsistency. \citet{yu2022consistency} use a similar consistency-based training which considers both box regression and classification in detection tasks at the same time.
\citet{kim2021lada} use Mixup~\cite{zhang2018mixup} to generate augmented samples during training and propose an acquisition function which looks ahead the effect of data augmentation. As data augmentation has exhibited great potentials in semi-supervised training, future active learning works can integrate with more powerful data augmentation strategies to alleviate the expensive data annotation, and improve the performance of training with insufficient data.

\section{Vision Tasks Empowered by DeepAL}
\label{tasks}

Thanks to its capability of selecting the most valuable samples and saving annotation costs, active learning has been widely used in deep model training for various vision tasks. This section proceeds to introduce active learning methods applied in different vision tasks. 

\subsection{Image Classification \& Recognition}
Image classification and recognition is the cornerstone of many vision tasks. The majority of studies on active learning in the vision community are conducted on this task.

Traditional active learning methods for image classification can be summarized into three categories: uncertainty-based approaches, diversity-based approaches, and expected model change, respectively. In the category of uncertainty-based approaches, LeastConf, Margin and Entropy are simple and effective ways to query informative samples. Also, distances to the decision boundary of the classifier can be used to evaluate the uncertainty~\cite{tong2001support,li2014multi}. Some other authors like~\citet{seung1992query,mccallumzy1998employing} train multiple models to construct a committee and measure the uncertainty by disagreement between different members of the committee. In the diversity-based approaches, \citet{nguyen2004active} cluster the data pool to select representative samples.~\cite{elhamifar2013convex,yang2015multi,guo2010active} optimize the data selection in a discrete space.~\cite{bilgic2009link,hasan2015context,mac2014hierarchical} consider the distance to surrounding data points and choose a subset to represent the global distribution. Expected model change aims to find samples that can cause substantial changes to the current model. It can be estimated from different perspectives, such as gradient length~\cite{settles2007multiple}, future errors~\cite{roy2001toward} and output changes~\cite{freytag2014selecting,kading2016active}, respectively.

Traditional methods introduced above cannot scale well to large-scale data and deep neural networks. Recent years have witnessed many active learning methods designed for deep learning. For example,~\cite{liu2016active,deng2018active,lin2018active} study the hyperspectral image classification for remote sensing.~\citet{liu2016active} train a DBN that includes two stages: unsupervised feature learning and supervised fine-tuning. The first stage can be used to estimate the representativeness, while the second stage is used to estimate the uncertainty. Medical image analysis requires medical experts for labeling, thus also faces the difficulty of obtaining well-annotated data. Many works attempt to introduce active learning to reduce the annotation cost~\cite{folmsbee2018active,bochinski2018deep,du2018breast,sayantan2018classification}. \citet{stark2015captcha} solve the CAPTCHA recognition by performing active learning. Their method does not require oracles to annotate a large training set. Once the classifier solves a CAPTCHA correctly, the label is automatically obtained. Since using all correctly classified data for re-training would be inefficient, only the most uncertain samples are selected for training.

\subsection{Object Detection}
Object detection requires the model to locate and classify objects simultaneously, thus it is more complicated than image classification. The demand for instance-level annotation imposes a higher annotation cost than labeling image categories, making active learning for object detection valuable. In recent years, active learning for object detection has attracted more attention. According to their motivations, those works can be divided into two categories: proposing specific query strategies for object detection and reducing annotation costs.

\subsubsection{Query Strategy}
Query strategy for object detection denotes the way of selecting samples for annotating object bounding boxes. Most existing works follow the setting that queries the whole image and annotates all objects in those images. Although various query strategies have been proposed in image classification to find valuable samples, these methods do not perform well when directly applied to object detection. One reason is that those methods make image-level queries based on instance-level predictions, which leads to inconsistency.

First, we introduce traditional active learning methods for object detection. \citet{yao2012interactive} use Hough forests as the detector, which detects objects using the generalized Hough transform and randomized decision trees. They estimate an optimal detection threshold based on the distribution of detection scores, which is hence used to model the annotation cost. Images with the highest annotation cost are selected because they are considered as most informative samples. \citet{bietti2012active} use a linear SVM in a sliding window as the detector, and query points closest to the decision boundary. \citet{roy2016active} model the object detection as a structured regression problem, and develop an active learning algorithm using the principled version space approach in the ``difference of feature'' space for structured prediction. The algorithm selects points for labeling such that the version space is maximally reduced.

 With the development of CNN, deep active learning for object detection has been extensively studied. The uncertainty-based methods are widely used for querying samples. \citet{feng2019deep} use MC dropout and Deep Ensembles to obtain uncertainty estimations and select the most uncertain samples by various score functions, \eg,\ entropy, mutual information. \citet{wang2018towards} estimate the uncertainty by performing the cross image validation, \ie, pasting proposals from the unlabeled image into a certain annotated image, then evaluating the prediction consistency under different image contexts. Region proposals with low consistency will be asked for annotation. \citet{roy2018deep} propose a query by committee paradigm based on the SSD detector. The disagreement for a particular candidate bounding box among different convolution layers is used for selecting samples. The intuition is that a detection network is likely to have similar predictions in a local region. Similarly, \citet{aghdam2019active} compute pixel-level uncertainty scores by the divergence among predictions at different levels of decoder for each pixel and their neighborhoods. Furthermore, the pixel-level scores are aggregated into an image-level score, which is used to select informative samples. \citet{brust2019active} evaluate a set of aggregation functions, including calculating the sum, the average and the maximum over all detection results. \citet{yuan2021multiple} argue that many methods ignore the interference from a large number of noisy instances in object detection, which leads to the inconsistency between instance-level and image-level uncertainty. Therefore, they introduce a multiple instance learning module to re-weight instance uncertainty. This method is expected to highlight the informative instances and depress noisy ones.

The above-mentioned methods mainly focus on the classification part in object detection for selecting samples. Some other works also consider the bounding box regression for selecting informative samples. \citet{choi2021active} learn a Gaussian Mixture Model (GMM) for the network outputs including both localization and classification. Consequently, both aleatoric and epistemic uncertainties are computed by parameters of the GMM, which include mean, variance and mixture weights. \citet{kao2018localization} introduce two metrics of localization uncertainty: Localization Tightness with the Classification Information (LT/C) and Localization Stability with the Classification Information (LS+C). LT/C is based on the overlap ratio between the region proposal and the final prediction. LS+C is based on the variation of predicted object locations when input images are corrupted by noises. \citet{yu2022consistency} apply data augmentations to unlabeled images, and measure the consistency between predictions of the original and augmented images. In contrast to LS+C, which considers box regression and classification independently, \citet{yu2022consistency} unify them with a single metric.

The diversity-based method has also been explored in object detection. Since the large receptive fields in deep CNNs may cause confusion among spatially neighboring classes, \citet{agarwal2020contextual} propose contextual diversity to measure the confusion associated with spatially co-occurring classes. Such a measure helps to select samples with diverse spatial contexts. Moreover, \citet{wu2022entropy} presents a hybrid method that considers both uncertainty and diversity. The uncertainty is estimated by basic detection entropy and the instance diversity can be divided into intra-image diversity and inter-image diversity. For the intra-image diversity, entropy-based Non-Maximum Suppression is performed to remove redundant instances within images. For the inter-image diversity, the class-specific prototypes of each image can help to improve the intra-class diversity, and the adaptive budget size for each class can enhance the inter-class diversity.

\subsubsection{Annotation Reduction}
In addition to the above works studying query strategies, there are other works aiming to further reduce the annotation cost. \citet{desai2019adaptive} propose an adaptive supervision framework that combines active learning with weakly supervised learning. All selected samples are first annotated with weak labels, \ie, center locations of objects. Those weak labels are then used to generate pseudo labels. Accurate bounding box annotations are required only for objects where the model can not produce confident predictions. The cost of different annotation types is measured by median annotation times in statistics. \citet{pardo2021baod} also consider a hybrid training set consisting of both weak and strong labels. Given a fixed budget, appropriate annotation action is chosen for each image based on certain measurements such as the uncertainty score. Then a teacher-student model is trained to make use of both types of annotations. Since the annotation cost of each image depends on the number of objects present in the image, \citet{desai2020towards} propose a fine-grained sampling strategy to selectively pick the most informative subset of bounding boxes rather than whole images. 

\subsection{Image Segmentation}

Image segmentation is also a fundamental task in computer vision with wide applications like scene understanding, medical image analysis, robotic perception, and autonomous driving. Image segmentation classifies each pixel in an image into a specific class. Annotating ground-truth for segmentation is hence more expensive than labeling image semantics and categories. Therefore, it is appealing to use active learning to relieve the annotation burden.

Existing active learning methods for image segmentation can be divided into image-based methods and region-based methods according to the granularity of selected data for annotation. Image-based approaches consider each image as a sample while region-based approaches divide unlabeled images into non-overlapping regions and consider each region as a sample. 

\subsubsection{Image-based Active Segmentation}
Image-based approaches for segmentation share the same intuition as standard pool-based batch-mode active learning.
Suggestive Annotation~\cite{yang2017suggestive} is a milestone work in deep active segmentation. It presents a new deep active learning framework by combining FCNs and active learning. Samples are first sorted by uncertainty and then the Core-set~\cite{sener2018active} is applied to select the batch for annotation. Instead of using another image descriptor network, it computes the overall uncertainty of each training sample with the mean of uncertainty of all pixels and calculates feature descriptors for the core-set by average pooling the output of the last convolution layer.
\citet{ozdemir2018active} propose a Borda-count based sample querying strategy, which considers both uncertainty and representativeness. Uncertainty is measured by the variance of predictions in their model. Representativeness is measured by ``Content Distance'', which is computed on the whole outputs of the last convolution layer before spatial pooling.
\citet{wang2018deep} propose a Nodule R-CNN model that attains state of-the-art pulmonary nodule segmentation performance. In addition, they train the model with a weakly-supervised method to leverage both labeled and unlabeled samples.

VAAL~\cite{sinha2019variational} proposes another framework for active image segmentation. It learns a VAE together with an adversarial network trained to discriminate unlabeled and labeled data. The proposed method implicitly learns the uncertainty for samples from the unlabeled pool. SRAAL~\cite{zhang2020state} designs an online uncertainty indicator, which endues unlabeled samples with different importance. Unlabeled samples with high confidence are treated as simple samples, and will not be selected for annotation. TA-VAAL~\cite{kim2021task} proposes to incorporate LL4AL and RankCGAN into VAAL by relaxing loss prediction with a ranker for ranking loss information.
\citet{agarwal2020contextual} introduce a novel information-theoretic distance measurement named Contextual Diversity (CD) to capture the diversity in spatial and semantic contexts of various object categories.

\subsubsection{Region-based Active Segmentation}
Different from image-based approaches, many active segmentation methods belong to region-based approaches. Region-based active learning is efficient since only a small number of regions is required to be labeled for each time.
In addition, since the Region of Interest (ROI) and confusing areas may only occupy a small percentage of the whole image, segmenting the whole image is not necessary. Those properties make region-based approaches perform better than image-based approaches under the same annotation budge.

One kind of region-based active segmentation methods divides the image into mesh.
\citet{mackowiak2018cereals} propose CEREALS, which only needs human to label a few automatically selected blocks.
\citet{casanova2020reinforced} propose a novel modification to the deep Q-network to select regions based on predictions and uncertainties.

Another kind of region-based active segmentation method utilizes superpixels to simplify the annotation. Superpixels are non-overlapping local regions generated by grouping similar pixels together. Object boundaries can be preserved among superpixels and pixels within each superpixel commonly share the same class. In this way, annotators only need to classify the selected superpixels instead of annotating each pixel on the image. 
Among those methods, \citet{siddiqui2020viewal} select samples according to viewpoint entropy and superpixel uncertainty.
\citet{equal2020} enforce pixel-wise self-consistency computed on outputs of segmentation network between each image and its transformation.
\citet{cai2021revisiting} propose a class-balanced score function to further boost the performance of the superpixel-based approach by favoring the selection of informative samples from under-represented object categories.
Another kind of works~\cite{Shin_2021_ICCV} design a labeling system which assists workers to label certain superpixels on the selected images. 

\subsection{Video Recognition}
Compared with image recognition, video recognition needs to consider both spatial cues and temporal correlations. Active learning for video recognition also shows great potentials to save the labeling cost.

\citet{vondrick2011video} propose an AL framework for object motion track prediction. The proposed framework selects frames with the largest difference in the predicted track of an object, if the frame is labeled. A large difference is generally caused by distractions like occlusions or cluttered backgrounds. This approach automatically labels tracks of stationary objects and leaves visually ambiguous tracks for manual annotation. \citet{hossain2018deactive} study human activity recognition and propose a deep and active learning enabled model (DeActive) that adopts a simple k-means clustering AL approach. DeActive clusters features and selects the most informative samples according to a density-weighted heuristic. Samples found within the most dense regions are generally more representative. 

Active learning has also been employed by person re-identification task, too. Person re-id matches query person images or tracklets across videos recorded by different cameras. An increasing demand for the accurate identification of persons from varying camera angles and times has attracted attention to person re-id. One fundamental challenge faced by person re-id is the difficulty of data annotation. Human annotators need to identify the same person across myriad non-overlapping cameras. This procedure is also easily affected by factors like varied camera viewpoints, lighting change, and similar-appearance imposters, making traditional supervised learning hard to generalize to real applications. Many works have thus proposed AL frameworks for person re-id. Those methods have achieved comparable performance to supervised learning while maintaining a substantially lower annotation cost.

Early AL approaches for person re-id maintains a gallery of labeled images for each person, and matches unlabeled images to those galleries. \citet{wang2016highly} follow this paradigm and propose a Highly Efficient Regression (HER) model for scalable person re-id. HER$^+$, an incremental version of the batch-based HER, is also introduced. A trichotomous joint exploration-exploitation active sampling criteria is applied. Three components of this criteria are 1) appearance diversity exploration that values unlabeled samples farthest away from labeled data, 2) matching uncertainty exploitation that values samples that do not closely resemble any galleries, and 3) ranking uncertainty exploitation that values samples resembling multiple galleries. These criteria balances uncertainty and diversity to select a representative subset of data.

Above methods for person re-id require an annotated gallery for each person, hence are still costly to some extent. Some other active learning methods do not require a labeled dataset for initialization, thus is gaining more popularity. Liu et al. \cite{liu2017early} propose an active learning framework with pairwise constraint (EALPC). EALPC enforces similar samples closer to each other, thereby facilitates the selection of a more representative set. Liu et al. \cite{liu2020pair} further introduce another active learning method with pairwise diversity maximization (EAL-PDM). Unlike EALPC, EAL-PDM selects the most diverse and informative pairs instead of instances for annotation. Pairwise uncertainty estimation and pairwise diversity maximization are used during the selection procedure.

Wang et al. \cite{wangm2018deep} also select pairs of data instead of instances to build AL algorithms. Specifically, they chose tracklets for annotation, which are successive bounding boxes of a tracked person. Their AL algorithm selects the most confident tracklets as true positives. This criterion is supported by their experimental results that, true positive matches are significantly more informative than other relations. Once tracklet pairs are chosen, an oracle is involved to confirm their relation and merge corresponding tracklet pairs. In addition, a view-aware approach is adopted to quickly identify true positives and filter false positives to enlighten the workload.

Liu et al. \cite{liu2019deep} leverage Reinforcement Learning (RL) in AL framework and introduce a Deep Reinforcement Active Learning (DRAL) method. In DRAL, the agent for AL sampling and the CNN for re-id are optimized independently. The agent has an AL policy that is updated iteratively based on hard triplet loss computed from binary pairwise relation produced by the oracle. As the AL policy is refined, the agent is able to select more informative samples that are then annotated and used to incrementally update the CNN.

In addition to the aforementioned approaches, a clustering approach like that used in \cite{hossain2018deactive} is also applied for person re-id. Gao et al. \cite{gao2021unsupervised} use a clustering approach to generate a base cluster to provide pseudo labels. Annotations for inter-centroid relations are then used to split and merge clusters to refine the clustering accuracy. The refined cluster then guides the refinement of the re-id CNN. Similarly, Jin et al. \cite{jin2022towards} propose Support Pair Active Learning (SPAL) that also starts with an unsupervised clustering. In SPAL, a dual certainty selection strategy searches for pairs with the highest likelihood of being a false negative or a false positive. After annotation, inter-support pair relations are used to derive Must-Link and Cannot-Link sets that propagate new labels to the rest of the cluster, as well as to update the re-id CNN.

Related works that we have discussed are summarized and categorized in Table \ref{tab:tab1}.

\section{Industrial Applications}
\label{industry}
This section commences by highlighting the importance of DeepAL in real applications. It then discusses about incorporating DeepAL into the industrial-level model production process, namely deep fusion of DeepAL with GUI-based model production software.

\subsection{Importance of DeepAL in Industry}
For a typical AI model manufacturing process in industry, the initial steps are to assign annotators to label data and construct a training set. An intuitive way is to label all available samples from the unlabeled data pool. This is not sustainable and scalable for industry, since it is prohibitive to label millions of samples.

A more appealing alternative is to form an iterative loop where each step randomly labels a subset of samples to train the model. In practical vision applications, most of collected datasets exhibit a long-tailed distribution. In other words, many head categories occupy the majority of samples, while lots of tail categories have a small number of samples. The consequence of random sample selection is that most of the selected samples are from several dominant categories and samples in tail categories are scarcely selected and annotated. This largely degrades the performance of the trained model. 

DeepAL has the potential to alleviate the long-tailed distribution issue in the raw dataset by selecting the most informative samples, hence significantly reducing labeling costs. Equipped with DeepAL, both the quality of the training set and model performance can be enhanced by the iterative annotation and training loop. Moreover, a closed loop of the AutoML systems can be established by integrating DeepAL and incremental model enhancement algorithms to greatly expand the model production efficiency. One possible solution to easily use DeepAL in the industry is to cope with DeepAL with GUI-based model production software.

\subsection{Coping DeepAL with GUI-based Model Production Software}
This part starts by discussing the necessity of building GUI-based model production software. GUI-based AI model production software mostly supports the full cycle of the AI model development, including data preparation, model design, training, and deployment. As active learning is closely related to data preparation, data annotation, and model training, it can be easily incorporated into those systems to decrease the cost of data annotation and accelerate the model development. DeepAL thus has shown great potentials to be applied in GUI-based production software.

When developing AI-based industrial applications, the process of interaction between customers and AI companies is usually tedious and time-consuming. For example, some customers may not have clear goals for the AI-empowered products they want to develop, and thus the algorithm experts need to interact repetitively with users to bring things to light. In addition, customers are often hesitant to invest heavily in a product until they see how effective it is. This could make for a slow and low-quality development process.

To tackle the challenge and put the initiative in the hands of users, GUI-based model production software has emerged to help public sector and enterprise customers build AI-empowered solutions for their data science problems \cite{das2020amazon, baylor2017tfx}. Typically, the software supports the full life cycle of a model development process. The implementation details of the software components, such as data preprocessing and model training, are usually hidden. Users can implement their ideas with just clicks of a mouse. The appearance of such easy-to-use software fundamentally advances the democratization of AI technologies and realizes the transformation of model production from fragmentation to scale.

There are many well-known software developed by tech giants or AI startups. We divide these software into two categories, depending on whether they integrate DeepAL component into them. Next, we firstly introduce some representative non-DeepAL-based model production software, such as SageMaker Studio\footnote{https://aws.amazon.com/sagemaker/data-scientist}, ModelArts\footnote{https://www.huaweicloud.com/intl/en-us/product/modelarts.html}, and EasyDL\footnote{https://ai.baidu.com/easydl/}, to help readers understand the specific functionalities supported by the modern software. Then we elaborate YMIR\footnote{http://www.viesc.com, https://github.com/IndustryEssentials/ymir}, a DeepAL-based software that can greatly facilitate the production of models at scale. 

\subsubsection{SageMaker Studio}
\begin{figure}[ht]
  \centering
  \includegraphics[width=\linewidth]{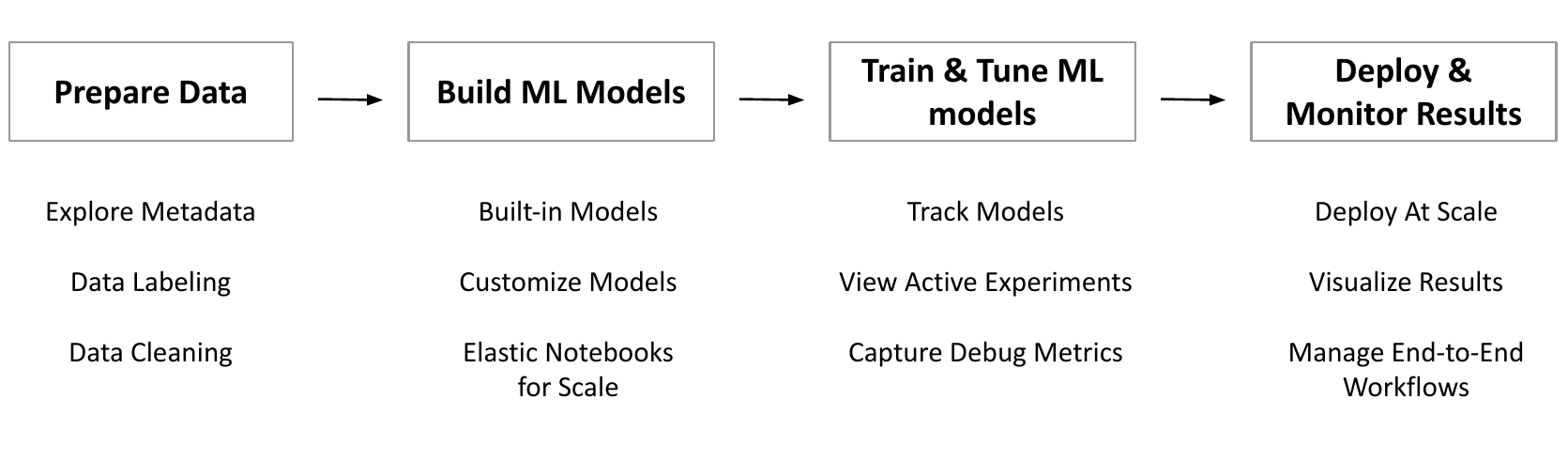}
  \caption{The main components of Amazon SageMaker Studio.}
  \label{fig:sagemaker}
\end{figure}

SageMaker Studio, developed by Amazon, provides a single, web-based visual interface for data science team to perform all machine learning steps. As shown in Figure \ref{fig:sagemaker}, it consists of four main components. During data preparation, users are allowed to prepare and label both structured and unstructured data at large scale. Afterwards, SageMaker not only provides all the tools to build machine learning models, but also relieves users of the need to manage infrastructure, allowing the models to be trained and tuned at scale. After users obtain satisfactory models, SageMaker makes it easy to deploy models and monitors the prediction results. 

\subsubsection{ModelArts}
ModelArts, developed by Huawei, is a one-stop GUI-based AI model production software that enables developers and data scientists of any skill level to quickly build, train, and deploy models anywhere, from the cloud to the edge. In addition to the basic components in machine learning life cycle, ModelArts offers unique features such as automatic labeling, neural architecture search, built-in large pretrained models, large-scale distributed training, and support for diverse training paradigms such as deep learning, reinforcement learning, and federated learning.

\subsubsection{EasyDL}
EasyDL is an AI development software developed by Baidu. It highlights two unique properties. First, the design of EasyDL is simple and extremely easy to understand, which is suitable for enterprise users who have zero algorithm foundation in AI. Second, EasyDL integrates Baidu Wenxin super large-scale pretraining model and self-developed AutoDL technology, so it can train a high-precision model based on a small amount of data.

\begin{figure*}[!t]
  \centering
  \includegraphics[width=1\textwidth]{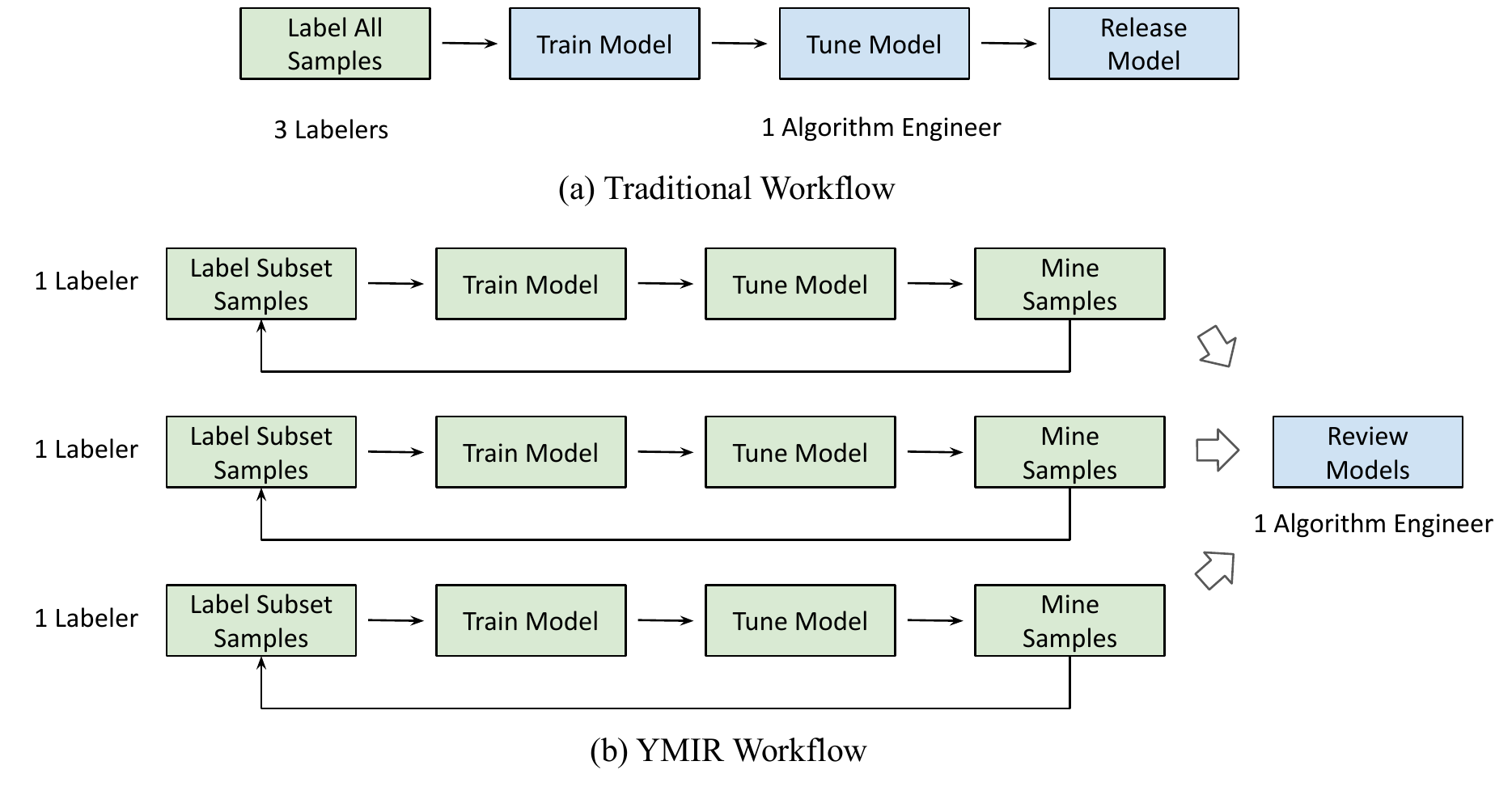}
  \caption{Comparisons between traditional model production workflow and YMIR model production workflow. The green and blue colors stand for the responsibilities for the annotator and the algorithm engineer, respectively. Here we assume there are three labelers and one algorithm engineer. In (a), labelers are assigned to annotate all samples in one task, and the algorithm engineer is responsible for the following steps. While in (b), each labeler is able to produce a model independently. They even can work on different tasks. The algorithm engineer only needs to review the produced models in this workflow.}
  \label{fig:workflow}
\end{figure*}

\subsubsection{YMIR}
Although the above software are powerful, they currently do not support DeepAL and users have to randomly select data for annotation, resulting in poor performance in actual use. Fortunately, \citet{huang2021ymir} introduces an open source GUI-based and DeepAL-based model production software You Mine In Recursion (YMIR) to enable the swift development of computer vision applications. The following part briefly reviews the pipeline of YMIR, and highlight its differences with previous model production software. More detailed descriptions to YMIR can be found in \citet{huang2021ymir}.

YMIR is an implementation of data-centric AI, which puts more emphasize on the quality of data procurement/mining instead of model optimization. It is the first AI algorithm development tool that integrate active learning methods with data version control in the process of developing AI algorithm. In addition, YMIR is completely code-free which lowers the technical requirements of its users to develop algorithms for industry application. In essence, YMIR is designed with the idea of easy development of AI algorithms for everybody while conforming to the industry standard and scale.

YMIR prioritizes the development of datasets in the product and system design. Therefore, the main components of YMIR are aligned with the typical steps in DeepAL, \ie, YMIR iteratively builds datasets and models through DeepAL mining, data labeling, and model training modules. It also borrows ideas from code version control to better track and manage data and models at different stages, and utilizes concepts such as projects to enable fast parallel iteration of multiple task-specific datasets. 
The included components are implemented through GUI interaction with explicit process guidance. Given a typical industrial example in the development of AI models, YMIR starts with few samples for the initialization of training data set, normally 100 samples for training, the validation will setup at the beginning of DeepAL circle and remain unchanged. The model start with a low accuracy (about 50\% of map50 in the validation set) YMIR gradually boost the model performance during the process of DeepAL, it starts with 100 samples for the first iteration and gather 500 samples during each iteration through DeepAL selection until the model performance match the requirement of the developer. In real industrial applications developers suffer from gathering enough training samples to start the AL model training due to the expensive labeling cost. YMIR introduces a new AI model production pipeline for user to start the model development with few samples and gain valuable data during each DeepAL iteration.  

The advent of YMIR changes the responsibilities of annotators and algorithm engineers in the model production process, as shown in Figure \ref{fig:workflow}. Traditionally, for a typical model production requirement in the industry, the algorithm engineer first hands over the obtained raw data to the annotators. Because the magnitude of data is usually in the thousands or tens of thousands, it takes the annotators lots of time to label all the data. After the annotation is completed, the algorithm engineer is responsible for training, tuning, and releasing the model. In contrast, by using YMIR, the annotators are able to independently produce models thanks to the GUI interface. More importantly, the integration of DeepAL in YMIR forms a cycle that includes data labeling, model training, model tuning, and DeepAL-based sample mining. In such a workflow, the algorithm engineer only needs to review the produced models and can concentrate on designing more advanced model architectures. The benefits of using DeepAL are twofold, namely a remarkably reduction in labeling efforts and a significant increase in model performance with similar manpower.

\section{Discussion of Future Directions}
\label{direction}
The effectiveness of DeepAL in computer vision has been demonstrated through extensive efforts. However, there are still many open problems to be coped with. We present six potential future directions in this section, the first three of which are more related to the training strategy of DeepAL and the latter three are more relevant to DeepAL algorithms.

\subsection{Universal Benchmark for DeepAL}
After reviewing the literature, we find that the reported performance of the random sampling baseline and some DeepAL methods differ significantly across works. For example, experimental evaluations in \citet{yu2022consistency} show that the method proposed in \citet{kao2018localization} is inferior to the random selection baseline, which contradicts with results presented in \citet{kao2018localization}.

Moreover, another issue is that different studies apply different settings when conducting experiments, making their performance not directly comparable. For instance, \citet{yoo2019learning} and \citet{kao2018localization} select different number of samples in the initial training sets. \citet{yuan2021multiple} and \citet{yu2022consistency} use different stop criteria, resulting in a different number of training samples in total. It is obvious that training with more data helps the final performance, which makes the improvements reported in the papers indefinite.

The above problems point to an urgent need to develop a fair performance assessment platform for determining a standard accuracy for the random sampling baseline and evaluating various DeepAL approaches. Fortunately, some researchers have noticed this issue and developed a benchmark named ALBench \cite{fengalbench}. In ALBench, different DeepAL algorithms share the same training and testing settings, effectively ensuring equitable comparisons. We believe ALBench will greatly facilitate the research in the community. While effective, ALBench is now only for object detection tasks. A further study may conduct to incorporate more computer vision tasks and the latest DeepAL algorithms.

\subsection{Integrating SSL with DeepAL}
Self-supervised learning (SSL) becomes very popular in these two years. It obtains supervisory signals from the tremendous unlabeled data itself and aims to learn generalizable and robust representations. The mainstream SSL approaches fall into the categories of contrastive-based and generative-based methods. Some well-known contrastive learning methods include SimCLR \cite{chen2020simple}, MoCo \cite{he2020momentum}, and BYOL \cite{grill2020bootstrap}. More recently, generative-based methods, \eg, MAE \cite{he2022masked}, BEiT \cite{bao2021beit}, and MST \cite{li2021mst}, has become the most successful self-supervised methods in the vision community and they have surpassed the promising performance achieved by contrastive learning methods.

A potential future direction is to combine SSL with DeepAL in order to further reduce the efforts of human labeling. The reason is that after pretraining by SSL methods, the model itself can already generalize well on downstream tasks. At this time, selecting few samples by DeepAL may be enough for enabling model to achieve competitive performance. However, \citet{chan2021marginal} have conducted extensive experiments on integrating SSL with DeepAL and found that the enhancement from DeepAL is marginal. They further performed ablation study to explore the reason. The results showed that data augmentation could be the source of this phenomena, as augmentation itself has strong capability in improving label efficiency. \citet{bengar2021reducing} also attempt to investigate whether SSL and DeepAL can be combined in a complementary way and they find that DeepAL can be beneficial for SSL only when the budget size is high. Therefore, the benefits of integrating SSL with DeepAL remain an open problem that deserves further exploration.

\subsection{Incremental Training in DeepAL}
Generally, existing methods require training a new model from scratch at each iteration of the DeepAL process. This training paradigm partially offsets the advantages of DeepAL, since retraining large-scale deep learning models is computationally expensive and time-consuming.

In order to tackle this issue and speed up model training, intuitively, we can combine the newly annotated samples with the initial training set and then fine-tune the model from the previous round \cite{shen2017deep}. However, some researchers have pointed out that such method introduces bias and degrades the model performance \cite{ostapuk2019activelink}. Specifically, the model tends to fit the new data and overwrites the learned knowledge. ActiveLink \cite{ostapuk2019activelink} delivers a unbiased incremental training approach to alleviate this problem and thus offers a balance between the newly and old annotation samples. Since this method is targeting for the link prediction in graphs, future research can be conducted to adapt ActiveLink or develop novel methods for vision tasks.

\subsection{Task Independent DeepAL}
Most DeepAL methods only target one or part of the subtasks in the computer vision domain. This trend is becoming more pronounced as researchers expect better performance on a certain subtask. For example in the object detection task, MI-AOD \cite{yuan2021multiple} argue that there is a gap between instance-level uncertainty and image-level uncertainty due to noisy instances in the background. They propose to apply a multiple instance learning module to force these two levels of uncertainty to be consistent.

The disadvantage of this fragmented situation is that the DeepAL algorithm is overly focused on specific subtasks and lacks generalization ability. Then a natural question is whether a unified DeepAL algorithm framework can be used to solve all mainstream computer vision tasks. This expectation of a unified algorithm is consistent with the current pursuit of unifying computer vision and natural language processing by using the same model architecture.

\subsection{DeepAL with Real-world Data}
Existing DeepAL methods are generally tested on crafted benchmark datasets, \eg, MS COCO and Pascal VOC in object detection, which are clean and balanced. However, they may fall short in dealing with complicated data distribution in real-world applications, \eg, out-of-distribution (OOD) samples and imbalanced data distribution. In industry, for example, we can get a lot of data from surveillance cameras. By looking closely at the data, we see that pedestrians, motor vehicles, and non-motor vehicles are present most of the time, while other objects to be detected are rarely present. This results in the collected datasets often exhibiting long-tailed distributions. Therefore, it is crucial to make DeepAL methods robust to these factors.

There has been some pioneering work to concentrate on this issue. For example, Contrastive Coding Active Learning (CCAL) \cite{du2021contrastive} point out that existing DeepAL methods often assume that labeled and unlabeled data come from the same class distribution. This assumption, however, does not hold when the unlabeled datasets contain many OOD samples. To address this, CCAL utilizes contrastive learning to extract both semantic and distinctive features and combines them in the query strategy to choose the unlabeled samples with matched categories. Another notable effort is Submodular Information Measures Based Active Learning (SIMILAR) \cite{kothawade2021similar}, which works to develop a unified DeepAL framework that takes care of many realistic scenarios such as OOD samples, rare classes, redundancy, and imbalanced datasets. Research in this direction is still at an early stage, and we hope that more research will focus on this direction in the future.

\subsection{Hierarchical Annotation for DeepAL}
Annotating the selected samples is an important step in the DeepAL process. Since DeepAL typically involves only a small fraction of data, the quality of the annotations determines the upper bound on the performance the models can achieve.

In some applications that require domain knowledge to label data, such as medical image analysis, the cost of hiring domain experts can be high. One possible direction to save cost is that, in addition to just selecting samples, DeepAL approaches also provide recommendations of the expertise required by samples, thus building a hierarchical annotation system.

For example, some brain tumor magnetic resonance imaging (MRI) labels can be difficult for junior doctors to judge, but easier for doctors with decades of experience. Moreover, the cost of hiring doctors of different specialties varies widely. If the DeepAL methods can provide the required expertise of the selected samples, we can save costs by only providing difficult samples to experienced doctors and easy samples to junior doctors.

\subsection{Industrial Integration of DeepAL}
Development Operations (DevOps) is a popular technique in the field of software engineering, whose purpose is to enable better communication and collaboration between developers and operations staff, and to deliver applications and services at high velocity and reliability through automated processes. With the advent of the AI era, the concept of MLOps has inevitably emerged. Similar to DevOps, MLOps unifies machine learning systems development and deployment, so as to standardize the model production process and enable continuous delivery of high-performance production models.

Many MLOps systems have been proposed for public use. For example, Amazon SageMaker MLOps\footnote{https://aws.amazon.com/cn/sagemaker/mlops/} helps machine learning engineers easily train, test, troubleshoot, deploy, and govern machine learning models at scale to increase productivity while maintaining model performance in production. Current systems have not integrated with DeepAL components. Since data-centric AI has attracted increasing attention and DeepAL is a key technology for enabling data iteration, integrating DeepAL into the MLOps workflow is an important direction for the future industrial application.

\section{Conclusion}
\label{conclude}
In this paper, we present a comprehensive review on DeepAL applied in computer vision. We first give a brief overview of AL and systematically summarize the recent developments in DeepAL. Then we describe the applications of DeepAL in common computer vision tasks, including image classification and recognition, object detection, image segmentation and video recognition, and show how DeepAL is adopted to industrial-level software and systems to accelerate model production. Lastly, we discuss the limitations of DeepAL and propose interesting directions and challenges for the future development of DeepAL.

\begin{table*}[t]
\caption{Taxology of the described DeepAL methods}
\centering
\tabcolsep=0.2mm
\begin{tabular}{l|ccc|cccc}
                               & \rotatebox{45}{Uncertainty} & \rotatebox{45}{Representative} & \rotatebox{45}{Semi supervised} & \rotatebox{45}{Classification} & \rotatebox{45}{Detection} & \rotatebox{45}{Segmentation} & \rotatebox{45}{Video} \\ \hline
\citet{gal2017deep}& $\checkmark$&&&$\checkmark$& $\checkmark$& $\checkmark$&\\
\citet{beluch2018power}& $\checkmark$&&& $\checkmark$& $\checkmark$&$\checkmark$&\\
\citet{sener2018active}&&$\checkmark$&&$\checkmark$&$\checkmark$&$\checkmark$&\\
\citet{yoo2019learning}&$\checkmark$&&&$\checkmark$&$\checkmark$&$\checkmark$&\\
\citet{sinha2019variational}&&$\checkmark$&$\checkmark$&$\checkmark$&$\checkmark$&$\checkmark$&\\
\citet{ash2019deep}&$\checkmark$&$\checkmark$&&$\checkmark$&&\\
\citet{kirsch2019batchbald}&$\checkmark$&$\checkmark$&&$\checkmark$&&\\
\citet{wang2016cost}&$\checkmark$&&$\checkmark$& $\checkmark$&&&\\
\citet{simeoni2021rethinking}           & $\checkmark$                &                                &   $\checkmark$                               & $\checkmark$                   &             &             &        \\
\citet{liu2016active}           & $\checkmark$                &                                &   $\checkmark$                               & $\checkmark$                   &             &             &        \\
\citet{gao2020consistency}          & $\checkmark$                &                                &   $\checkmark$                               & $\checkmark$                   &             &             &        \\
\citet{kim2021lada}         & $\checkmark$                &                                &   $\checkmark$                               & $\checkmark$                   &             &             &        \\
\citet{kao2018localization}    & $\checkmark$                &                                &                                 &                                & $\checkmark$              &                              &                       \\
\citet{roy2018deep}            & $\checkmark$                &                                &                                 &                                & $\checkmark$              &                              &                       \\
\citet{desai2019adaptive}      & $\checkmark$                &                                &                                 &                                & $\checkmark$              &                              &                       \\
\citet{brust2019active}        & $\checkmark$                &                                &                                 &                                & $\checkmark$              &                              &                       \\
\citet{agarwal2020contextual}  &                             & $\checkmark$                   &                                 &                                & $\checkmark$              &                              &                       \\
\citet{yuan2021multiple}       & $\checkmark$                &                                & $\checkmark$                    &                                & $\checkmark$              &                              &                       \\
\citet{choi2021active}         & $\checkmark$                &                                &                                 &                                & $\checkmark$              &                              &                       \\
\citet{yu2022consistency}      & $\checkmark$                & $\checkmark$                   & $\checkmark$                    &                                & $\checkmark$              &                              &                       \\
\citet{wu2022entropy}          & $\checkmark$                & $\checkmark$                   &                                 &                                & $\checkmark$              &                              &                       \\ 
\citet{gorriz2017cost}           & $\checkmark$                &                                &   $\checkmark$                               &                   &             &   $\checkmark$           &        \\
\citet{yang2017suggestive}     & $\checkmark$                & $\checkmark$                   &                                 &                                &                           & $\checkmark$                 &                       \\
\citet{ozdemir2018active}      & $\checkmark$                & $\checkmark$                   &                                 &                                &                           & $\checkmark$                 &                       \\
\citet{wang2018deep}           & $\checkmark$                & $\checkmark$                   & $\checkmark$                    &                                &                           & $\checkmark$                 &                       \\
\citet{agarwal2020contextual}  & $\checkmark$                & $\checkmark$                   &                                 &                                &                           & $\checkmark$                 &                       \\
\citet{mackowiak2018cereals}   & $\checkmark$                &                                & $\checkmark$                    &                                &                           & $\checkmark$                 &                       \\
\citet{casanova2020reinforced} & $\checkmark$                &                                & $\checkmark$                    &                                &                           & $\checkmark$                 &                       \\
\citet{siddiqui2020viewal}     & $\checkmark$                &                                & $\checkmark$                    &                                &                           & $\checkmark$                 &                       \\
\citet{equal2020}              & $\checkmark$                &                                & $\checkmark$                    &                                &                           & $\checkmark$                 &                       \\
\citet{cai2021revisiting}      & $\checkmark$                & $\checkmark$                   & $\checkmark$                    &                                &                           & $\checkmark$                 &                       \\ 
\citet{Shin_2021_ICCV}         & $\checkmark$                & $\checkmark$                   & $\checkmark$                    &                                &                           & $\checkmark$                 &                       \\
\citet{liu2017early}           &                             & $\checkmark$                   &                                 &                                &                           &                              & $\checkmark$          \\
\citet{hossain2018deactive}    & $\checkmark$           & $\checkmark$                   &         &                  &              &           & $\checkmark$          \\
\citet{liu2019deep}            & $\checkmark$                &                                &                                 &                                &                           &                              & $\checkmark$          \\
\citet{liu2020pair}            & $\checkmark$                & $\checkmark$                   &                                 &                                &                           &                              & $\checkmark$          \\
\citet{gao2021unsupervised}    & $\checkmark$                & $\checkmark$                   & $\checkmark$                    &                                &                           &                              & $\checkmark$          \\
\citet{jin2022towards}         & $\checkmark$                &                                & $\checkmark$                    &                                &                           &                              & $\checkmark$         
\end{tabular}
\label{tab:tab1}
\end{table*}




\section*{Statement of interest}
None.

\vskip2pc
\bibliography{refs}





\end{document}